\documentclass[runningheads]{llncs}

\usepackage[T1]{fontenc}
\usepackage{graphicx}
\usepackage{amsmath,amssymb}
\usepackage{booktabs}
\usepackage{multirow}
\usepackage{hyperref}
\usepackage{xspace}
\usepackage[table, dvipsnames]{xcolor}
\usepackage{tikz}
\usetikzlibrary{arrows.meta,positioning,fit,calc,shapes.geometric,backgrounds}
\usepackage{pgfplots}
\usepackage{subcaption}
\usepackage{placeins}
\pgfplotsset{compat=1.18}
\hypersetup{hidelinks}

\newcommand{\method}{HemaHier\xspace}
\newcommand{\datasetDELPHI}{UKA2\xspace}
\newcommand{\datasetUK}{UKA1\xspace}
\newcommand{\datasetMLL}{MLLv1\xspace}

\begin{document}

\title{HemaHier: Chain-Conditioned Ordinal Hierarchies for Lineage-Aware Bone-Marrow Cytology}

\author{
    Afshin Bozorgpour\inst{1} \and
    Peter Schüffler\inst{2,3,4} \and
    Edgar Jost\inst{5,6} \and
    Dorit Merhof\inst{1,7}
}

\institute{
Faculty of Informatics and Data Science, University of Regensburg, Germany\\
\and 
Institute of Pathology, TUM School of Medicine and Health, Technical University of Munich, Germany
\and
Munich Center for Machine Learning, Germany
\and
Munich Data Science Institute, Technical University of Munich, Germany
\and 
Department of Hematology, Oncology, Hemostaseology and Stem Cell Transplantation, Medical School, RWTH Aachen University, Aachen, Germany\\
\and
Center for Integrated Oncology Aachen Bonn Cologne Düsseldorf, Aachen, Germany
\and 
Fraunhofer Institute for Digital Medicine MEVIS, Bremen, Germany
\\
\email{dorit.merhof@ur.de}
}

\authorrunning{A. Bozorgpour et al.}
\titlerunning{HemaHier: Lineage-Aware Bone-Marrow Cytology}

\maketitle

\begin{abstract}
Bone-marrow cytology is inherently structured: each cell belongs to a hematopoietic lineage, and many cell types lie on ordered maturation trajectories. Standard flat classifiers ignore this structure, treating a mild same-lineage confusion the same as a severe cross-lineage mistake and predicting only discrete labels. We propose \method{}, an ordinal-hierarchical prediction head for a frozen or lightly adapted cytology foundation model. Its central component is a chain-conditioned maturity score that reads a single maturity value under a per-chain query, supervised only on biologically valid healthy chains, while dysplastic and off-chain cell types remain classes but are excluded from maturity supervision. Fine and lineage predictions are coupled through a shared posterior that guarantees hierarchical consistency, and a staged objective first stabilizes recognition, then adds lineage and maturity supervision. 
On three bone-marrow datasets under a shared ontology, \method{} achieves competitive recognition while reducing biologically severe errors and adding a within-lineage maturity ordering that flat classifiers lack.
Code is available at \url{https://github.com/xmindflow/HemaHier}.

\end{abstract}

\section{Introduction}
\label{sec:intro}

Bone-marrow cytology is central to the diagnosis and monitoring of hematologic disease. 
In routine assessment, hematologists inspect marrow aspirate smears and interpret both single-cell morphology and differential cell counts. 
Automated single-cell recognition has therefore become an important target for computational hematology, with deep learning demonstrating strong performance on peripheral-blood and bone-marrow cell classification~\cite{acevedo2020pbc,choi2017white,matek2021highly}, including efforts to make recognition robust across imaging domains and laboratories~\cite{sadafi2023continual}.
Despite this progress, most models still treat recognition as flat multi-class classification, which is particularly limiting for rare but clinically relevant identities, where small errors in recognition or lineage assignment can distort downstream differential counts and interpretation~\cite{tayebi2022automated,ghete2024models}.

This flat formulation ignores a key property of marrow cytology: the label space is biologically structured. 
Each cell belongs to a hematopoietic lineage, and many healthy identities occupy ordered positions along a maturation trajectory~\cite{laurenti2018haematopoietic}. 
A promyelocyte confused with a myelocyte is a nearby same-lineage maturation error, whereas a promyelocyte confused with an erythroblast is a cross-lineage error. 
Standard cross-entropy treats both as equally wrong, and common metrics such as top-1 accuracy or macro-F1 do not reveal whether mistakes are biologically mild or severe. 
For clinical and biological interpretation, however, the severity of an error matters: preserving lineage consistency and maturation order can be as important as maximizing fine-label accuracy alone.

The emergence of hematology foundation models further changes the problem.
Self-supervised models such as DinoBloom provide strong transferable embeddings for single-cell blood and bone-marrow images~\cite{koch2024dinobloom,oquab2024dinov2}, and interpretability methods built on them expose morphological concepts to experts~\cite{dasdelen2025cytosae}, shifting the question from whether a backbone can separate cell types to whether the prediction head exposes the structure hematologists rely on: lineage, maturation, and error severity.
Hierarchical classification encourages semantically or biologically better mistakes~\cite{deng2010whatdoesclassify,bertinetto2020making}, and guided and ordinal representation learning arrange hematopoietic cells along expert-defined lineage and maturity relations~\cite{graebel2021guided,graebel2022spatial,cao2020coral}; closest to our setting, biologically-informed backbone adaptation penalizes dangerous cross-lineage confusions on marrow cytology~\cite{muminov2025cytodino}.
These approaches, however, model lineage and maturation primarily through structured representations or lineage-specific trajectories; none couples a lineage-consistent fine posterior with a continuous, chain-specific maturity read-out while explicitly separating healthy maturation chains from atypical, dysplastic, or off-chain identities.
Maturation may not be faithfully represented by a single global scalar alone, because different hematopoietic lineages follow distinct visual trajectories, and dysplastic or ambiguous cells may belong to a lineage yet lack a well-defined maturity position.


We propose \method{}, an ordinal-hierarchical prediction head for bone-marrow cytology. It augments a frozen or lightly adapted cytology foundation model with factorized lineage and detail representations, coupled fine and lineage classifiers, and a chain-conditioned maturity head; maturity is supervised only on biologically valid healthy chains, while atypical and off-chain identities remain in the label space but are excluded from maturity losses. 
Our contributions are threefold: (i) a shared bone-marrow ontology separating fine identity, lineage, healthy maturation chains, and off-chain variants; (ii) a chain-conditioned maturity score that yields a continuous maturity ordering for each biologically valid maturation chain through a shared query-conditioned head; and (iii) an evaluation that reports not only recognition but also error severity and maturation quality, testing prediction plausibility.

\FloatBarrier
\section{Method}
\label{sec:method}

\begin{figure}[!ht]
    \centering
    \resizebox{0.9\linewidth}{!}{%
    \begin{tikzpicture}[
    font=\scriptsize,
    >=Latex,
    box/.style={draw, rounded corners=2pt, align=center, inner sep=3pt, minimum height=8mm},
    bluebox/.style={box, fill=blue!7, draw=blue!45!black},
    greenbox/.style={box, fill=green!8, draw=green!45!black},
    orangebox/.style={box, fill=orange!10, draw=orange!55!black},
    purplebox/.style={box, fill=purple!8, draw=purple!45!black},
    graybox/.style={box, fill=gray!10, draw=gray!55!black},
    arrow/.style={->, thick},
    dashedarrow/.style={->, thick, dashed, gray!80!black}
    ]
    \node[bluebox, text width=20mm] (ont) at (12.7,1.4)
    {\textbf{Ontology}\\
    fine class $f$\\
    lineage $m(f)$\\
    chains $\mathcal{C}_q$\\
    off-chain kept};
    
    \node[draw=blue!45!black, rounded corners=2pt, inner sep=1.2pt, fill=blue!7] (x) at (0,1.55)
    {\includegraphics[width=11mm]{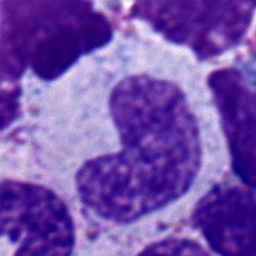}};
    \node[font=\scriptsize, align=center] at (0,0.7) {cell crop\\$x$};
    
    \node[bluebox, text width=20mm] (backbone) at (2.1,1.4)
    {cytology\\backbone $g_\theta$\\
    frozen / LoRA};
    
    \node[purplebox, text width=26mm] (rep) at (5.6,1.4)
    {\textbf{factorized features}\\
    $z_c=\phi_c(h_L)$\\
    $z_d=\phi_d([h_L;h_M])$};
    
    \node[orangebox, text width=22mm] (heads) at (8.7,1.4)
    {\textbf{coupled heads}\\
    $p_f(f\mid x)$\\
    $p_c(c\mid x)$\\
    $s_q(x)\in[0,1]$};
    
    \draw[arrow] (x) -- (backbone);
    \draw[arrow] (backbone) -- node[above, font=\tiny] {$h_L,h_M$} (rep);
    \draw[arrow] (rep) -- (heads);
    
    \node[graybox, text width=28mm] (decode) at (5.0,-0.3)
    {\textbf{hierarchy decoding}\\
    argmax\\
    Bayesian product};
    
    \node[greenbox, text width=28mm] (maturity) at (8.7,-0.3)
    {\textbf{ordinal readout}\\
    maturity score\\
    ranking metrics\\
    tree-distance analysis};
    
    \draw[arrow] (heads.south west) -- (decode);
    \draw[arrow] (heads.south) -- (maturity.north);
    
    \draw[dashedarrow] (ont.north) -- (12.7, 2.5) -|
    node[above, font=\tiny, align=center] {fine + lineage labels} (heads.north);
    
    \draw[dashedarrow] (ont.west) -- node[above, font=\tiny, align=center, rotate=38] {valid healthy\\chains only} (maturity.north east);
    
    \node[graybox, text width=26mm] (offchain) at (12.4,-0.3)
    {atypical / dysplastic\\off-chain identities\\
    no maturity loss};
    
    \draw[dashedarrow] (ont.south) -- (offchain);
    \draw[dashedarrow] (offchain) -- (maturity);
    
    \end{tikzpicture}%
    }
    \caption{\textbf{Overview of \method{}.} Bone-marrow labels are represented by fine classes, lineages, valid healthy maturation chains, and off-chain identities. A cytology backbone provides late and intermediate features, which are factorized into lineage context $z_c$ and detail/maturity representation $z_d$. The coupled heads predict fine identity, lineage, and branch-conditioned maturity. Maturity supervision is applied only to valid healthy chains, while hierarchy-conditioned decoding combines fine and lineage evidence at inference.}
    \label{fig:method_overview}
    
    \end{figure}
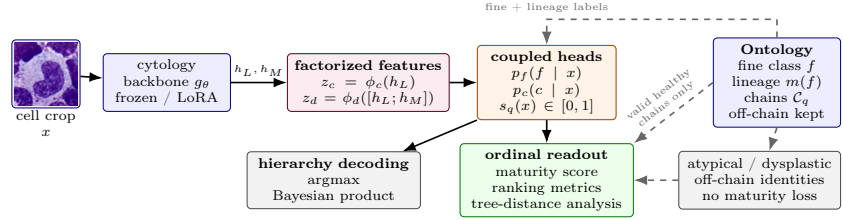

\method{} is an ordinal-hierarchical prediction head built on top of a cytology backbone. 
The key idea is to decouple three related but distinct factors of bone-marrow interpretation: fine cell identity, hematopoietic lineage, and within-lineage maturity. 
Figure~\ref{fig:method_overview} summarizes the ontology, factorized representation, chain-conditioned maturity heads, and hierarchy-conditioned inference.

\vspace{-0.1cm}
\paragraph{Ontology and valid maturity chains.}
Each fine cell type $f$ belongs to exactly one hematopoietic lineage, given by a fixed expert-defined mapping $m(f)$. 
Within a lineage, the healthy cell types follow a known maturation order, which we encode as a set of expert-curated chains
\begin{equation}
\mathcal{C}_q=(u_{q,0},\ldots,u_{q,n_q}), \qquad q\in\mathcal{Q},
\end{equation}
where $u_{q,j}$ is the $j$-th fine class of chain $q$, ordered from immature ($j{=}0$) to mature ($j{=}n_q$).
A cell of type $u_{q,j}$ then receives a normalized maturity target $\pi=j/n_q\in[0,1]$.
Dysplastic, atypical, and other off-chain cell types keep their lineage but lie on no chain, so they stay valid fine classes yet receive no maturity target.
This avoids forcing weakly ordered or abnormal cells onto a healthy developmental axis.
The same ontology also defines a maturity-aware \emph{tree distance} (TD) between a predicted and a true label, used only to measure error severity, never as a training loss; it is the TD metric reported in Table~\ref{tab:main}. A same-chain confusion costs the normalized maturity gap between the two stages, a same-lineage off-chain confusion an intermediate constant, and a cross-lineage confusion the maximum, so the metric ranks biologically mild maturation slips below severe lineage-inconsistent errors.

\paragraph{Factorized representation.}
A backbone $g_\theta$ yields a late semantic token $h_L$ and an intermediate morphology-preserving feature $h_M$. Two adapters produce a lineage context $z_c=\phi_c(h_L)$ and a detail representation $z_d=\phi_d([h_L;h_M])$ that captures chromatin texture, granularity, and within-lineage maturity cues; the fine representation is their concatenation $[z_c;z_d]$.

\paragraph{Coupled heads and consistent posterior.}
Rather than predicting fine class and lineage independently, HemaHier couples them into a single lineage-consistent posterior. A lineage head gives a distribution $p_c(c \mid x)$ over lineages, and a fine head predicts a raw distribution $\tilde{p}_f(f \mid x)$ over fine classes. We retain only the fine probability assigned to the predicted lineage $m(f)$ and renormalize it within that lineage:
\begin{equation}
p_f(f \mid x)=
\underbrace{p_c(m(f)\mid x)}_{\text{lineage probability}}
\cdot
\underbrace{
\frac{\tilde{p}_f(f\mid x)}
{\sum_{f':\,m(f')=m(f)} \tilde{p}_f(f'\mid x)}
}_{\text{fine class within its lineage}},
\label{eq:posterior}
\end{equation}
which corresponds to the factorization
$p_f(f\mid x)=p_f(f\mid m(f),x)\,p_c(m(f)\mid x)$
and guarantees that fine predictions remain consistent with their predicted lineage. The raw head $\tilde{p}_f$ is trained directly with a lightly weighted auxiliary cross-entropy, which stabilizes early optimization. Maturity is predicted as a single chain-conditioned score $s_q(x)=g(z_c,z_d,e_q)$, read under a learned chain embedding $e_q$, so all chains share one maturity head selected by the chain query rather than a separate regressor per chain.

\paragraph{Training.}
With lineage labels $y_i^c=m(y_i^f)$ and, for on-chain samples $i\in\Omega$, chain $q_i$, predicted maturity $s_i=s_{q_i}(x_i)$, and normalized stage target $\pi_i$, we minimize
\begin{equation}
\mathcal{L}=\mathcal{L}_{\mathrm{joint}}+\lambda_{\mathrm{raw}}\mathcal{L}_{\mathrm{raw}}+\lambda_c(t)\,\mathcal{L}_{c}+\lambda_m(t)\,\mathcal{L}_{m}+\lambda_r(t)\,\mathcal{L}_{r},
\label{eq:loss}
\end{equation}
combining the negative log-likelihood of the posterior in Eq.~\eqref{eq:posterior}, the auxiliary fine cross-entropy, and a lineage classifier with two maturity terms on on-chain samples: a regression of each score to its stage target, and a margin ranking that keeps a later stage above an earlier one within the same chain,
\begin{equation}
\mathcal{L}_m=\frac{1}{|\Omega|}\sum_{i\in\Omega}\big(s_i-\pi_i\big)^2,
\qquad
\mathcal{L}_r=\!\!\sum_{\substack{q_i=q_j\\ \pi_i<\pi_j}}\!\!\max\!\big(0,\;\delta-(s_j-s_i)\big),
\label{eq:maturity}
\end{equation}
where $\delta$ is the ranking margin. The structural weights $\lambda(t)$ ramp in sequence so fine discrimination stabilizes before lineage, maturity, and ranking.

\paragraph{Inference.}
At test time we decode each cell from the hierarchy-consistent posterior $p_f(f\mid x)$, which by construction keeps the predicted fine class and its lineage in agreement. Optionally, this posterior can be recalibrated by the chain-conditioned maturity score, favouring fine classes whose expected maturation position matches the predicted one. Models are selected on validation macro-F1 under the hierarchy-consistent posterior.

\FloatBarrier
\section{Experimental results}
\label{sec:results}
\subsection{Setup}

\textbf{Datasets and ontology.}
We evaluate on two in-house bone-marrow cytology datasets, \datasetDELPHI{} (9,118 cells, 23 fine classes) and \datasetUK{} (8,566 cells, 24 fine classes), and the public MLLv1 (BMC) bone-marrow reference~\cite{matek2021highly} (171,374 cells, 21 fine classes). Dataset-specific labels are harmonized into a shared ontology of fine identities, lineages, healthy maturation chains, and off-chain variants. Each fine class maps to one lineage, and maturity targets are defined only for biologically valid healthy chains; atypical, dysplastic, and ambiguous identities remain valid fine classes but are excluded from maturity supervision.
\\
\\
\noindent
\textbf{Protocol and baselines.} 
We use five-fold stratified cross-validation per dataset, sharing splits, preprocessing, augmentation, sampler, optimizer, and training budget across all methods. Unless noted, the backbone is DinoBloom-S~\cite{koch2024dinobloom}, a compact hematology foundation model; our method is backbone-agnostic, and we refer to it simply as the backbone below. A shared frozen backbone isolates the effect of the prediction head; we additionally report a LoRA-adapted~\cite{hu2022lora} (rank 16) backbone. Because public bone-marrow splits differ across works, all methods are re-trained and evaluated under our common five-fold protocol, and checkpoints are selected by validation macro-F1.
We compare against a flat linear probe, the standard foundation-model evaluation protocol, and HMIL$^{*}$, our single-cell adaptation of the hierarchical multi-instance framework of Jin et al.~\cite{jin2025hmil}: the original aggregates a bag of patches by attention~\cite{ilse2018attention} for whole-slide images, which degenerates when each image is a single cell, so we retain its coarse/fine hierarchical head and train it on the same features as every method. The ablation additionally includes a multi-scale flat variant that reads the same features without the hierarchy.
\\
\\
\noindent
\textbf{Implementation and metrics.}
Models use $224\times224$ crops with the backbone's standard preprocessing, AdamW (weight decay $10^{-4}$), batch size 64, 20 epochs, a cosine schedule with linear warmup, and learning rate $3\times10^{-4}$; the adapted setting uses LoRA~\cite{hu2022lora} rank 16. The lineage, maturity, and ranking losses use weights $\lambda_c{=}0.05$, $\lambda_m{=}0.005$, and $\lambda_r{=}0.003$, ramped in after warmup, and the auxiliary fine cross-entropy weight is increased under adaptation. On top of the frozen backbone ($21.6$M parameters), the \method{} head adds $5.0$M trainable parameters, and under adaptation a rank-16 LoRA adds a further $0.44$M; the backbone is never fully fine-tuned, and the chain-conditioned maturity components that produce the ordinal axis account for under $0.25$M of the added parameters. Experiments ran on NVIDIA TITAN RTX GPUs. For recognition we report accuracy, balanced accuracy, fine and rare-class macro-F1, and weighted F1. For error severity we use the cross-lineage error rate (CLE) and the maturity-aware tree distance (TD) from Sec.~\ref{sec:method}, and we assess maturation quality by the Spearman correlation and pairwise order accuracy between predicted and reference positions on healthy on-chain cells.
\subsection{Results}
We evaluate \method{} against a flat linear probe and the HMIL$^{*}$ method on three bone-marrow cytology datasets under a frozen backbone, where every method shares identical features and only the prediction head differs, and a lightly LoRA-adapted backbone (Table~\ref{tab:main}). We report standard recognition metrics together with error-severity and maturity-aware metrics.
\begin{table}[!ht]
  \centering
  \caption{\textbf{Main comparison (5-fold mean$\pm$std).} CLE: cross-lineage error rate; TD: maturity-aware tree distance; Mat-$\rho$: within-chain maturity rank correlation, which only \method{} predicts. HMIL$^{*}$ is our single-cell adaptation of~\cite{jin2025hmil}. Best per column in bold; \method{} shaded.}
  \label{tab:main}
  \resizebox{\linewidth}{!}{%
    \begin{tabular}{ll|ccccc|cc|c}
\toprule
Dataset & Method & Acc$\uparrow$ & B-Acc$\uparrow$ & M-F1$\uparrow$ & W-F1$\uparrow$ & Rare$\uparrow$ & CLE$\downarrow$ & TD$\downarrow$ & Mat-$\rho$ \\
\midrule
\multicolumn{10}{c}{\it Frozen backbone} \\
\midrule
\multirow{3}{*}{\datasetUK{}} & Flat & 72.1\,{\footnotesize$\pm$0.9} & 65.2\,{\footnotesize$\pm$0.7} & 61.2\,{\footnotesize$\pm$1.6} & 73.6\,{\footnotesize$\pm$0.8} & 54.4\,{\footnotesize$\pm$3.9} & 10.0\,{\footnotesize$\pm$0.4} & 15.4\,{\footnotesize$\pm$0.5} & -- \\
 & HMIL$^{*}$ & 71.9\,{\footnotesize$\pm$1.6} & \textbf{67.7\,{\footnotesize$\pm$1.2}} & \textbf{61.8\,{\footnotesize$\pm$1.8}} & 73.5\,{\footnotesize$\pm$1.4} & \textbf{55.2\,{\footnotesize$\pm$6.7}} & 9.6\,{\footnotesize$\pm$0.4} & 15.0\,{\footnotesize$\pm$0.7} & -- \\
\rowcolor{orange!8}
 & \method{} & \textbf{74.0\,{\footnotesize$\pm$1.6}} & 63.4\,{\footnotesize$\pm$1.9} & 61.7\,{\footnotesize$\pm$3.2} & \textbf{74.5\,{\footnotesize$\pm$1.6}} & 54.1\,{\footnotesize$\pm$9.9} & \textbf{9.5\,{\footnotesize$\pm$0.4}} & \textbf{14.2\,{\footnotesize$\pm$0.6}} & 0.871 \\
\midrule
\multirow{3}{*}{\datasetDELPHI{}} & Flat & 88.4\,{\footnotesize$\pm$0.9} & \textbf{87.9\,{\footnotesize$\pm$1.2}} & 86.9\,{\footnotesize$\pm$0.7} & 88.5\,{\footnotesize$\pm$0.9} & 84.1\,{\footnotesize$\pm$1.2} & 3.6\,{\footnotesize$\pm$0.5} & 5.7\,{\footnotesize$\pm$0.6} & -- \\
 & HMIL$^{*}$ & 87.1\,{\footnotesize$\pm$0.7} & 86.4\,{\footnotesize$\pm$0.5} & 85.1\,{\footnotesize$\pm$0.6} & 87.2\,{\footnotesize$\pm$0.7} & 81.1\,{\footnotesize$\pm$1.4} & 3.9\,{\footnotesize$\pm$0.4} & 6.3\,{\footnotesize$\pm$0.4} & -- \\
\rowcolor{orange!8}
 & \method{} & \textbf{89.5\,{\footnotesize$\pm$0.5}} & 87.7\,{\footnotesize$\pm$0.7} & \textbf{87.6\,{\footnotesize$\pm$0.6}} & \textbf{89.5\,{\footnotesize$\pm$0.5}} & \textbf{84.3\,{\footnotesize$\pm$2.9}} & \textbf{3.1\,{\footnotesize$\pm$0.3}} & \textbf{4.9\,{\footnotesize$\pm$0.3}} & 0.887 \\
\midrule
\multirow{3}{*}{\datasetMLL{}} & Flat & 81.5\,{\footnotesize$\pm$0.4} & \textbf{76.5\,{\footnotesize$\pm$1.5}} & 63.3\,{\footnotesize$\pm$1.6} & 82.8\,{\footnotesize$\pm$0.3} & 38.6\,{\footnotesize$\pm$5.4} & 12.7\,{\footnotesize$\pm$0.5} & 16.1\,{\footnotesize$\pm$0.5} & -- \\
 & HMIL$^{*}$ & 78.1\,{\footnotesize$\pm$0.5} & 75.9\,{\footnotesize$\pm$1.4} & 55.0\,{\footnotesize$\pm$0.6} & 80.1\,{\footnotesize$\pm$0.4} & 17.1\,{\footnotesize$\pm$1.5} & 14.9\,{\footnotesize$\pm$0.4} & 18.9\,{\footnotesize$\pm$0.4} & -- \\
\rowcolor{orange!8}
 & \method{} & \textbf{84.2\,{\footnotesize$\pm$0.4}} & 75.7\,{\footnotesize$\pm$1.3} & \textbf{72.2\,{\footnotesize$\pm$0.9}} & \textbf{84.8\,{\footnotesize$\pm$0.3}} & \textbf{59.1\,{\footnotesize$\pm$3.1}} & \textbf{10.6\,{\footnotesize$\pm$0.3}} & \textbf{13.7\,{\footnotesize$\pm$0.3}} & 0.947 \\
\midrule
\multicolumn{10}{c}{\it LoRA-adapted backbone} \\
\midrule
\multirow{3}{*}{\datasetUK{}} & Flat & 77.7\,{\footnotesize$\pm$0.8} & 67.3\,{\footnotesize$\pm$1.6} & 66.4\,{\footnotesize$\pm$2.1} & 78.0\,{\footnotesize$\pm$0.8} & 59.1\,{\footnotesize$\pm$5.3} & 8.3\,{\footnotesize$\pm$0.4} & 12.2\,{\footnotesize$\pm$0.5} & -- \\
 & HMIL$^{*}$ & 76.9\,{\footnotesize$\pm$1.4} & \textbf{69.7\,{\footnotesize$\pm$2.5}} & 66.3\,{\footnotesize$\pm$3.0} & 77.8\,{\footnotesize$\pm$1.3} & 58.9\,{\footnotesize$\pm$8.3} & 8.4\,{\footnotesize$\pm$0.5} & 12.7\,{\footnotesize$\pm$0.7} & -- \\
\rowcolor{orange!8}
 & \method{} & \textbf{78.0\,{\footnotesize$\pm$0.7}} & 67.6\,{\footnotesize$\pm$2.2} & \textbf{66.9\,{\footnotesize$\pm$2.9}} & \textbf{78.2\,{\footnotesize$\pm$0.6}} & \textbf{59.7\,{\footnotesize$\pm$10.2}} & \textbf{8.2\,{\footnotesize$\pm$0.6}} & \textbf{12.0\,{\footnotesize$\pm$0.5}} & 0.891 \\
\midrule
\multirow{3}{*}{\datasetDELPHI{}} & Flat & 91.2\,{\footnotesize$\pm$0.6} & 90.3\,{\footnotesize$\pm$0.8} & 89.9\,{\footnotesize$\pm$0.5} & 91.2\,{\footnotesize$\pm$0.6} & 88.4\,{\footnotesize$\pm$2.4} & 2.4\,{\footnotesize$\pm$0.1} & 4.0\,{\footnotesize$\pm$0.3} & -- \\
 & HMIL$^{*}$ & 90.4\,{\footnotesize$\pm$0.6} & 89.8\,{\footnotesize$\pm$0.5} & 89.3\,{\footnotesize$\pm$0.2} & 90.4\,{\footnotesize$\pm$0.6} & 87.7\,{\footnotesize$\pm$1.6} & 2.7\,{\footnotesize$\pm$0.3} & 4.4\,{\footnotesize$\pm$0.3} & -- \\
\rowcolor{orange!8}
 & \method{} & \textbf{91.7\,{\footnotesize$\pm$0.6}} & \textbf{90.6\,{\footnotesize$\pm$0.4}} & \textbf{90.6\,{\footnotesize$\pm$0.2}} & \textbf{91.7\,{\footnotesize$\pm$0.6}} & \textbf{89.4\,{\footnotesize$\pm$2.4}} & \textbf{2.2\,{\footnotesize$\pm$0.4}} & \textbf{3.8\,{\footnotesize$\pm$0.4}} & 0.916 \\
\midrule
\multirow{3}{*}{\datasetMLL{}} & Flat & 86.9\,{\footnotesize$\pm$0.4} & 79.2\,{\footnotesize$\pm$1.3} & 76.3\,{\footnotesize$\pm$1.4} & 87.4\,{\footnotesize$\pm$0.3} & 65.0\,{\footnotesize$\pm$5.3} & 8.8\,{\footnotesize$\pm$0.3} & 11.4\,{\footnotesize$\pm$0.3} & -- \\
 & HMIL$^{*}$ & 85.9\,{\footnotesize$\pm$0.2} & \textbf{79.8\,{\footnotesize$\pm$2.0}} & 74.2\,{\footnotesize$\pm$2.2} & 86.4\,{\footnotesize$\pm$0.2} & 60.9\,{\footnotesize$\pm$7.5} & 9.6\,{\footnotesize$\pm$0.2} & 12.3\,{\footnotesize$\pm$0.2} & -- \\
\rowcolor{orange!8}
 & \method{} & \textbf{87.4\,{\footnotesize$\pm$0.3}} & 79.1\,{\footnotesize$\pm$1.8} & \textbf{77.9\,{\footnotesize$\pm$1.7}} & \textbf{87.7\,{\footnotesize$\pm$0.3}} & \textbf{69.2\,{\footnotesize$\pm$6.2}} & \textbf{8.4\,{\footnotesize$\pm$0.3}} & \textbf{11.0\,{\footnotesize$\pm$0.3}} & 0.959 \\
\bottomrule
\end{tabular}
  }
\end{table}
\\
\\
\noindent
\textbf{Frozen backbone}. 
On the frozen backbone every method reads identical features, so differences primarily reflect the prediction head. \method{} attains the best fine and rare-class F1 on \datasetDELPHI{} and \datasetMLL{}, by far the largest margin on the rare-class-imbalanced \datasetMLL{}, where it improves fine macro-F1 by $8.9$ points ($63.3\!\to\!72.2$) and rare-class F1 by over twenty points ($38.6\!\to\!59.1$) over the strongest baseline; the \datasetMLL{} recognition gains and the cross-lineage-error reductions on all three datasets are significant (paired $t$-test over folds, $p<0.05$). On \datasetUK{} the methods are tied on macro-F1 (within the cross-fold deviation), as several classes contribute only one or two cells per fold; \method{} still leads on the more stable accuracy, weighted F1, and both severity metrics, and adds a maturity read-out a flat classifier cannot express.
\\
\\
\noindent
\textbf{LoRA adaptation}.
With a rank-16 adapter and a strengthened auxiliary fine cross-entropy that keeps adaptation recognition-driven, \method{} attains the best accuracy and fine and rare-class F1 on all three datasets (e.g.\ $+1.6$ macro-F1 and $+4.2$ rare-class F1 over the adapted flat probe on \datasetMLL{}) together with the lowest cross-lineage and tree-distance errors, so the structural advantages persist under adaptation. The margin narrows on the pathology-heavy \datasetUK{}, where one lineage dominates and many cells are blasts, dysplastic, or off-chain, so the lineage prior has little to disambiguate; there \method{} still leads the flat probe and HMIL$^{*}$ on accuracy, weighted F1, macro-F1, and both severity metrics, and uniquely provides a calibrated maturation ordering ($\rho{=}0.89$).

\begin{figure*}[!ht]
  \centering
  \includegraphics[width=0.8\linewidth]{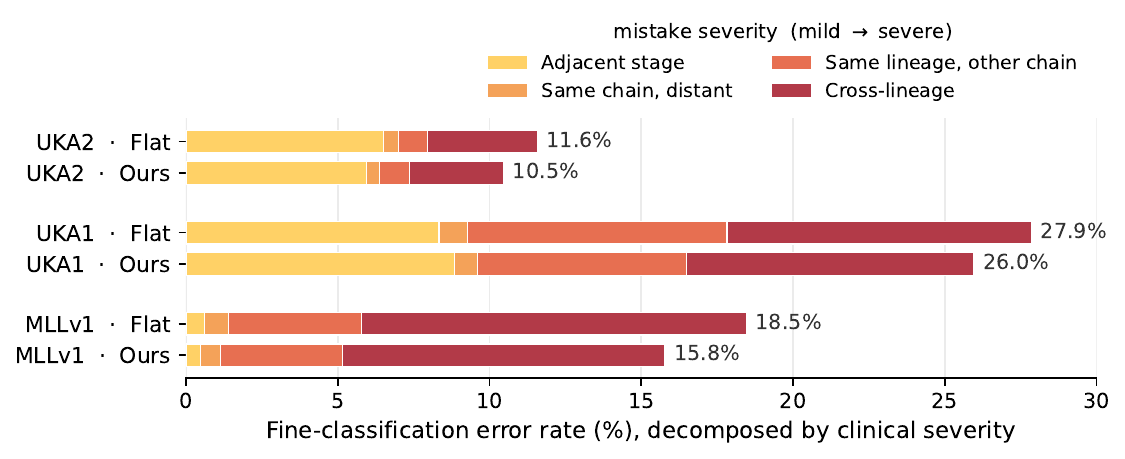}
  \caption{\textbf{Errors are fewer and milder} (frozen backbone, 5-fold mean). Each bar is the fine-classification error rate decomposed by clinical severity, from an adjacent-stage maturation slip (mildest) to a cross-lineage confusion (severest). On every dataset \method{} both shortens the bar (fewer errors) and shrinks its severe (dark) portion relative to the flat probe, most visibly the cross-lineage errors on \datasetMLL{} and the off-chain confusions on \datasetUK{}.}
  \label{fig:severity}
\end{figure*}

\paragraph{Error severity.}
Clinically, how a model errs matters: a same-lineage maturation slip is far milder than a cross-lineage confusion. Decomposing errors by severity (Fig.~\ref{fig:severity}), \method{} makes fewer and milder mistakes, with the lowest cross-lineage error and tree distance across datasets.

\begin{figure}[!ht]
  \centering
  \includegraphics[width=\linewidth]{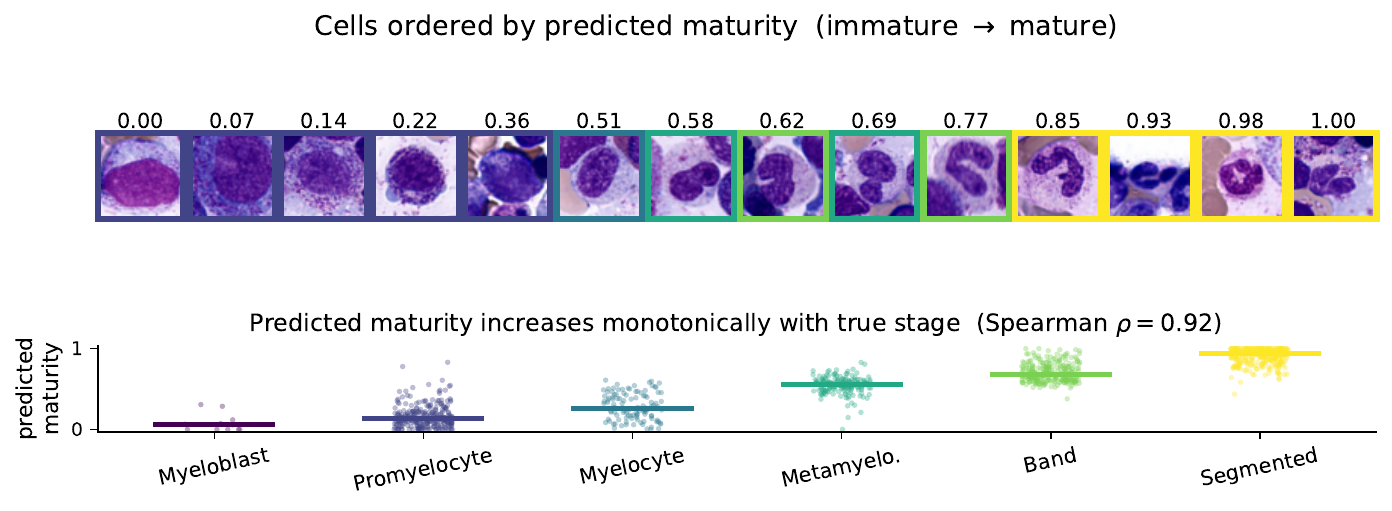}
  \caption{\textbf{Learned maturation axis.} \emph{Top:} real \datasetUK{} neutrophil-lineage cells ordered left-to-right by \method{}'s predicted maturity (score shown above each; border colour is the true stage). \emph{Bottom:} predicted maturity versus true stage over all on-chain cells; the score rises monotonically along the chain.}
  \label{fig:maturity}
\end{figure}

\paragraph{Maturity structure.}
The chain-conditioned maturity head recovers a continuous within-chain ordering rather than only discrete labels. Figure~\ref{fig:maturity} shows representative \datasetUK{} neutrophil-lineage cells ordered by predicted maturity, which rises monotonically with true stage (Spearman $\rho=0.92$, on-chain cells only). This ordering is consistent across chains ($\rho=0.87$--$0.95$, Table~\ref{tab:main}) and cannot be produced by a flat classifier.

\subsection{Component ablation}
Table~\ref{tab:ablation} adds components one at a time on \datasetDELPHI{}, whose macro-F1 is near-saturated so the value is clearest on severity. Factorization, the lineage-consistent posterior, and chain-conditioned maturity progressively lower cross-lineage error and tree distance ($3.6\!\to\!3.0$ and $5.7\!\to\!4.9$), and within-chain ranking yields a small gain in recognition, although the maturity and severity metrics do not further improve, indicating a trade-off between fine-class discrimination and ordinal structure. At inference, the hierarchy-consistent Bayesian product decoder preserves recognition while further lowering both severity metrics without retraining; we report argmax for recognition.
\begin{table}[!ht]
  \centering
  \scriptsize
  \caption{\textbf{Ablation and decoding on \datasetDELPHI{} (5-fold mean$\pm$std).} Frozen backbone, validation-argmax selection; components are added in turn and \method{} is the full model.}
  \label{tab:ablation}
  \setlength{\tabcolsep}{3pt}
\begin{tabular}{l|cc|cc|cc}
\toprule
Configuration & M-F1\,$\uparrow$ & Rare\,$\uparrow$ & CLE\,$\downarrow$ & TD\,$\downarrow$ & Mat.\,$\rho\uparrow$ & Ord.\,$\uparrow$ \\
\midrule
\multicolumn{7}{l}{\it Component ablation (frozen backbone; 5-fold mean$\pm$std)} \\
\midrule
Flat ($h_L$ only) & 86.9{$\pm$0.7} & 84.1{$\pm$1.2} & 3.6{$\pm$0.5} & 5.7{$\pm$0.6} & -- & -- \\
\ + Multi-scale ($h_L,h_M$) & 86.4{$\pm$0.5} & 82.9{$\pm$1.3} & 3.5{$\pm$0.1} & 5.6{$\pm$0.3} & -- & -- \\
\ + Factorized + lineage & 87.1{$\pm$0.5} & 83.6{$\pm$1.9} & 3.2{$\pm$0.4} & 5.1{$\pm$0.4} & -- & -- \\
\ + Branch maturity & 87.2{$\pm$1.1} & 83.4{$\pm$3.8} & 3.0{$\pm$0.3} & 4.9{$\pm$0.3} & 0.90 & 96.3{$\pm$0.6} \\
\ + Within-chain ranking (\method{}) & 87.5{$\pm$0.8} & 84.5{$\pm$2.6} & 3.1{$\pm$0.5} & 5.0{$\pm$0.4} & 0.89 & 92.2{$\pm$0.6} \\
\midrule
\multicolumn{7}{l}{\it Decoder analysis (same \method{} checkpoints; no retraining)} \\
\midrule
\method{} + argmax & 87.5{$\pm$0.8} & 84.5{$\pm$2.6} & 3.1{$\pm$0.5} & 5.0{$\pm$0.4} & 0.89 & 92.2{$\pm$0.6} \\
\method{} + Bayesian product & 87.6{$\pm$0.6} & 84.8{$\pm$2.3} & 3.0{$\pm$0.5} & 4.9{$\pm$0.4} & 0.89 & 92.2{$\pm$0.6} \\
\bottomrule
\end{tabular}
\end{table}
\FloatBarrier

\section{Conclusion}
\label{sec:conclusion}
\method{} is an ontology-guided prediction head that makes a frozen or lightly adapted cytology backbone lineage- and maturation-aware. Across three datasets, it achieves competitive recognition, with particularly strong rare-class performance on the most imbalanced dataset, while making residual errors that are biologically more plausible: lower cross-lineage error and tree distance at a fixed recognition budget. Beyond labels, it recovers a calibrated within-lineage maturity ordering that flat classifiers cannot express, and its hierarchy-consistent decoder can trade toward lineage safety at inference without retraining. Its main limitation is reliance on expert-curated maturation chains, a natural target for future extension to pathological trajectories.

\bibliographystyle{splncs04}
\bibliography{ref}

\end{document}